\documentclass[pdflatex,sn-basic,Numbered,iicol]{sn-jnl}

\usepackage{graphicx}%
\usepackage{amsmath,amssymb,amsfonts}%
\usepackage[table]{xcolor}%
\usepackage{textcomp}%
\usepackage{booktabs}%
\usepackage{bm}%

\newcommand{\gref}{g_{\theta}}
\newcommand{\gstar}{g_{\theta^{*}}}
\newcommand{\rres}{r_{\phi}}
\newcommand{\rori}{r_{\xi}}
\newcommand{\Dtr}{\mathcal{D}_{S}^{\mathrm{tr}}}
\newcommand{\Dval}{\mathcal{D}_{S}^{\mathrm{val}}}
\newcommand{\Dtgt}{\mathcal{D}_{T}}
\newcommand{\bce}{\ell}
\newcommand{\fwd}{HUI360\,$\rightarrow$\,\allowbreak SSUP-A}
\newcommand{\rev}{SSUP-A\,$\rightarrow$\,\allowbreak HUI360}
\newcommand{\dAP}{\Delta\mathrm{AP}}
\definecolor{ourtint}{HTML}{FBEDE3}

\AtBeginDocument{\renewcommand{\tnote}[1]{\textsuperscript{#1}}}
\makeatletter
\newcommand{\tabnote}[2]{\par{\removelastskip\vskip3pt\tablefootnotefont\noindent\if!#1!\else\textsuperscript{#1}\fi#2\par}}
\makeatother

\newcommand{\APAreaFwd}{0.178}
\newcommand{\APFFwd}{0.129}
\newcommand{\APFRev}{0.321}
\newcommand{\APGFwd}{0.129}
\newcommand{\APGRev}{0.277}
\newcommand{\APHGBallRev}{0.427}
\newcommand{\APHGBposeFwd}{0.055}
\newcommand{\APQFwd}{0.048}
\newcommand{\APRFgeoRev}{0.369}
\newcommand{\APrTgtRev}{0.436}
\newcommand{\AUCgTgtFwd}{0.841}
\newcommand{\AUCgTgtRev}{0.733}
\newcommand{\AUCgValFwd}{0.923}
\newcommand{\AUCgValRev}{0.871}
\newcommand{\AUCrTgtFwd}{0.650}
\newcommand{\AUCrTgtRev}{0.765}
\newcommand{\AUCrValFwd}{0.911}
\newcommand{\AUCrValRev}{0.745}
\newcommand{\CIFAreaFwd}{[$-$0.065, $-$0.033]}
\newcommand{\CIFAreaRev}{[$-$0.153, $+$0.117]}
\newcommand{\CIFGFwd}{[$-$0.016, $+$0.018]}
\newcommand{\CIFGRev}{[$+$0.024, $+$0.082]}
\newcommand{\CIFHGBallRev}{[$-$0.253, $+$0.012]}
\newcommand{\CIFJwFwd}{[$-$0.028, $-$0.001]}
\newcommand{\CIFixConfGRev}{[$-$0.001, $+$0.0002]}
\newcommand{\CIFixFGRev}{[$+$0.027, $+$0.091]}
\newcommand{\CIFixFJwFwd}{[$-$0.013, $+$0.022]}
\newcommand{\CIFixFJwRev}{[$-$0.047, $+$0.041]}
\newcommand{\CIFixNoVelGRev}{[$+$0.009, $+$0.064]}
\newcommand{\CaseAR}{$+$0.982}
\newcommand{\DFAreaFwd}{$-$0.049}
\newcommand{\DFAreaRev}{$-$0.020}
\newcommand{\DFCFwd}{$-$0.016}
\newcommand{\DFCRev}{$-$0.005}
\newcommand{\DFGFwd}{$-$0.0003}
\newcommand{\DFGRev}{$+$0.044}
\newcommand{\DFHGBallRev}{$-$0.106}
\newcommand{\DFJFwd}{$-$0.009}
\newcommand{\DFJRev}{$-$0.005}
\newcommand{\DFJwFwd}{$-$0.014}
\newcommand{\DFixConfRev}{$+$0.048}
\newcommand{\DFixFGRev}{$+$0.048}
\newcommand{\DFixFJwFwd}{$+$0.005}
\newcommand{\DFixFJwRev}{$+$0.005}
\newcommand{\DFixNoVelGRev}{$+$0.028}
\newcommand{\FPermMaxRev}{$+$0.054}
\newcommand{\FPermMinRev}{$+$0.039}
\newcommand{\LodoMaxFwd}{$+$0.003}
\newcommand{\LodoMaxRev}{$+$0.054}
\newcommand{\LodoMinFwd}{$-$0.005}
\newcommand{\LodoMinRev}{$+$0.040}
\newcommand{\ORCIFHAreaFwd}{[$-$0.059, $-$0.020]}
\newcommand{\ORCIFHFFwd}{[$+$0.003, $+$0.020]}
\newcommand{\ORCIFHFRev}{[$+$0.001, $+$0.018]}
\newcommand{\ORCIFHHGBaqHRev}{[$-$0.316, $-$0.103]}
\newcommand{\ORDFHAreaFwd}{$-$0.038}
\newcommand{\ORDFHFFwd}{$+$0.011}
\newcommand{\ORDFHFRev}{$+$0.007}
\newcommand{\ORDFHHGBaqHRev}{$-$0.196}
\newcommand{\ORFHFBLowRev}{$-$0.0001}
\newcommand{\ORFaceAUCMax}{0.86}
\newcommand{\ORFaceAUCMin}{0.68}
\newcommand{\ORKeptFeatures}{12}
\newcommand{\ORParTrFH}{385}
\newcommand{\ORPitchSrcMaxFwd}{0.81}
\newcommand{\ORPitchSrcMinFwd}{0.75}
\newcommand{\ORPitchTgtMaxFwd}{0.38}
\newcommand{\ORPitchTgtMinFwd}{0.32}
\newcommand{\ORSeedsFHFFwd}{5/5}
\newcommand{\ORSeedsFHFRev}{4/5}
\newcommand{\ORrhAUCFwd}{0.729}
\newcommand{\ORrhAUCRev}{0.760}
\newcommand{\ORrhAbsFwd}{1.01}
\newcommand{\ORrhAbsRev}{0.19}
\newcommand{\ORrqAUCFwd}{0.650}
\newcommand{\ORrqAUCRev}{0.765}
\newcommand{\ORrqAbsFwd}{4.93}
\newcommand{\ORrqAbsRev}{0.89}
\newcommand{\PFAddedNeg}{1{,}128}
\newcommand{\PFCIFGFwd}{[$-$0.014, $+$0.019]}
\newcommand{\PFCIFGRev}{[$+$0.029, $+$0.100]}
\newcommand{\PFDFGFwd}{$+$0.003}
\newcommand{\PFDFGRev}{$+$0.051}
\newcommand{\PFLenAddHUISrc}{434}
\newcommand{\PFLenNegHUISrc}{136}
\newcommand{\PFLenPosHUISrc}{406}
\newcommand{\PFPositives}{564}
\newcommand{\PFSeedsFGRev}{5/5}
\newcommand{\PermGMaxRev}{$+$0.009}
\newcommand{\PermGMinRev}{$-$0.007}
\newcommand{\SCDCRev}{$+$0.015}
\newcommand{\SCDHFwd}{$+$0.039}
\newcommand{\SCDHRev}{$+$0.047}
\newcommand{\SCDatesFiveFwd}{9}
\newcommand{\SCDatesHRev}{4/5}
\newcommand{\SDgRev}{4.8}
\newcommand{\SDrRev}{1.1}
\newcommand{\SRAPSelFFwd}{0.160}
\newcommand{\SRAPSelFRev}{0.402}
\newcommand{\SRAPSelRefFwd}{0.149}
\newcommand{\SRAPSelRefRev}{0.316}
\newcommand{\SRBoot}{500}
\newcommand{\SRCIFHGBallFwd}{[$+$0.012, $+$0.047]}
\newcommand{\SRCIFHGBallRev}{[$-$0.163, $+$0.074]}
\newcommand{\SRCIFHGFRev}{[$-$0.091, $-$0.014]}
\newcommand{\SRCIFRefFwd}{[$-$0.008, $+$0.023]}
\newcommand{\SRCIFRefRev}{[$+$0.054, $+$0.138]}
\newcommand{\SRCIHGFHGRefFwd}{[$-$0.016, $+$0.005]}
\newcommand{\SRCIHGFHGRefRev}{[$+$0.046, $+$0.163]}
\newcommand{\SRDFHGBallFwd}{$+$0.031}
\newcommand{\SRDFHGBallRev}{$-$0.035}
\newcommand{\SRDFHGFRev}{$-$0.050}
\newcommand{\SRDFRefFwd}{$+$0.011}
\newcommand{\SRDFRefRev}{$+$0.086}
\newcommand{\SRDHGFHGRefFwd}{$-$0.006}
\newcommand{\SRDHGFHGRefRev}{$+$0.090}
\newcommand{\SRRecSelFFwd}{0.04}
\newcommand{\SRRecSelFRev}{0.12}
\newcommand{\SRSeedsFRefRev}{5/5}
\newcommand{\SRSelHistGBFwd}{3}
\newcommand{\SRSelHistGBRev}{7}
\newcommand{\SRSelMLPFwd}{8}
\newcommand{\SRSelMLPRev}{5}
\newcommand{\SRSelNFwd}{15}
\newcommand{\SRSelRFFwd}{4}
\newcommand{\SRSelRFRev}{3}
\newcommand{\TPCIHGBhAcvFwd}{[$-$0.002, $+$0.055]}
\newcommand{\TPCIHGBhAcvRev}{[$+$0.020, $+$0.185]}
\newcommand{\TPDHGBhAcvFwd}{$+$0.026}
\newcommand{\TPDHGBhAcvRev}{$+$0.098}
\newcommand{\TPHGBaqHAcvRev}{0.525}

\begin{document}

\title[Cross-dataset cue transfer in interaction anticipation]{Fixed-Reference Pose Residuals for Measuring Cross-Dataset Cue Transfer in Human--Robot Interaction Anticipation}

\author[1]{\fnm{Bowen} \sur{Yang}}\email{bowenyang@njust.edu.cn}

\author[1]{\fnm{Xinliang} \sur{Xiao}}\email{xinliangxiao@njust.edu.cn}

\author[1]{\fnm{Wenjing} \sur{Zhang}}\email{924110800235@njust.edu.cn}

\author[1]{\fnm{Li} \sur{Yang}}\email{yangli945@njust.edu.cn}

\author*[1]{\fnm{Wei} \sur{Zhou}}\email{weizhou@njust.edu.cn}

\affil*[1]{\orgdiv{School of Automation}, \orgname{Nanjing University of Science and Technology}, \orgaddress{\city{Nanjing}, \postcode{210094}, \country{China}}}

\abstract{Social and service robots in public spaces need to anticipate which nearby person is about to approach and touch them, so that a response can be prepared before contact. It is largely unknown which cues support this anticipation when a model trained with one robot is used on another robot at a different site. We study this question with a fixed-reference pose residual (FRPR) model: a geometry predictor built from the person's bounding box and mask is trained and frozen, and a temporal network then learns from body pose an additive correction to its logit, so that every prediction splits exactly into a geometry term and a pose term. Between two public egocentric datasets recorded by different robots, HUI360 and SSUP-A, with every choice made on source data, the pose correction raised average precision (AP) from \APGRev{} to \APFRev{} from SSUP-A to HUI360 and gave no measurable gain in the opposite direction; the same asymmetry held over a stronger, source-selected geometry reference. Freezing gave no AP advantage over joint training, and simple geometric baselines and tree ensembles remained competitive or better, so the construction serves measurement rather than prediction accuracy. A head-orientation residual added small gains in both directions. Post hoc, whether a person faces the camera kept its discriminative direction across datasets, whereas head pitch reversed. With thresholds chosen on source data, the neural models that use geometry detected at most 17\% of target interactions. Code and processed data are available at \url{https://github.com/WeiZhou96/FRPR-interaction-anticipation}.}


\maketitle

\section{Introduction}\label{sec:intro}

Service and delivery robots that share sidewalks, lobbies and shops with the public are approached, touched, stopped and handed objects by people they have never met \citep{bu2024ssup,lorenzolouis2026hui360,arreghini2024service}. Approach, greeting and engagement behaviours designed for such robots, and socially aware navigation among people, depend on an early estimate of who is going to interact \citep{satake2009approach,kato2015may,avelino2025greeting,mavrogiannis2023core,francis2025principles}. About one second of warning allows a robot to slow down, turn its interface towards the person or keep moving when nobody needs it. The two kinds of anticipation error have different costs. A missed interaction leaves the robot unprepared when a person reaches it; a false alarm makes it stop, turn or address someone who is only passing, which interrupts its task and may be perceived as intrusive \citep{riosmartinez2015proxemics,mead2017proxemics}. The lead time with which a correct warning arrives determines which of these behaviours can still be executed. Both kinds of error are borne by the people around the robot, which makes the estimate a concern of human-centred system design \citep{ozmengaribay2023six} and not only of recognition accuracy.

The perception problem has recently been cast as \emph{interaction anticipation}: given a short observation of a tracked person in the robot's egocentric video, predict whether a physical interaction with the robot follows \citep{abbate2024self,lorenzolouis2026hui360}. The label is an observable event, contact with the robot body, rather than a mental state, which makes the task well defined and lets it be annotated at scale. Three families of cue are available for each track. \emph{Approach geometry} describes where the person appears in the image, how large the box and mask are, and how these quantities change. \emph{Body pose} is the sequence of two-dimensional keypoints, which carries body orientation, arm movement and gait. \emph{Head orientation} indicates whether a person faces the robot, a signal that engagement models for social robots have long used as evidence of attention \citep{michalowski2006spatial,sidner2005engagement,anzalone2015engagement,belardinelli2024gaze}. Existing anticipation models feed such cues to a single classifier \citep{abbate2024self,arreghini2024predicting,lorenzolouis2026hui360}.

A social robot is rarely deployed where its training data were recorded. Robots differ in camera height and placement, and sites differ in layout, crowd density and the reasons people have to approach. Cross-dataset results show that anticipation performance drops when the platform changes \citep{lorenzolouis2026hui360}, and performance in intent prediction depends on the robot's environment \citep{thompson2024pard}. For a designer, the useful question is which cues keep their meaning under such a change. Geometry is low-dimensional and tied to the task by construction, since nobody touches a robot from afar; pose and head orientation depend on the estimator, the camera and the viewpoint, and pose estimates from robot-mounted cameras suffer from occlusion and scale variation in crowds \citep{vendrow2023jrdb}. Once all cues enter one fused network, the contribution of each can no longer be read from the model, and, as Section~\ref{sec:components} shows, the same holds for an additive two-branch model trained jointly.

We therefore treat the geometry predictor as a \emph{reference} and ask each further cue for a \emph{correction}. A small geometry network is trained on the source domain and frozen; a causal temporal convolutional network then learns from the pose sequence a scalar that is added to the reference logit, starting from a zero-initialised head. We call this the fixed-reference pose residual (FRPR) model. The reference remains a deployable predictor with unchanged weights, every prediction splits exactly into a reference logit and a pose correction, and the difference between FRPR and its own reference is a paired measurement of what pose adds under that reference. A further cue is attached in the same way as one more frozen stage; with head orientation we obtain FRPR-H. The construction is intended as a measurement instrument, not as a stronger predictor, and we evaluate it as such.

We study both transfer directions between HUI360 \citep{lorenzolouis2026hui360} and SSUP-A, the annotation of the SSUP-HRI street recordings \citep{bu2024ssup} released with HUI360. All preprocessing statistics, hyperparameters, checkpoints and thresholds are chosen on source data. From SSUP-A to HUI360 the pose correction raises average precision (AP) over its reference by \DFGRev{} (95\% interval \CIFGRev), and controls with permuted or coordinate-free pose input show no such gain; from HUI360 to SSUP-A no gain is observed. The asymmetry persists over a stronger reference selected on source data, so the pose gain is not only a compensation for a weak reference. Freezing brings no AP advantage over joint training, and simple geometric baselines and tree ensembles are competitive or better. Geometry and pose keep their ranking quality in opposite transfer directions; among head-orientation cues, whether a person faces the camera keeps its discriminative direction across the two robots and sites, whereas head pitch reverses. At thresholds chosen on source data, the neural models that use geometry detect at most 17\% of target interactions.

The contributions are as follows.
\begin{itemize}
\item A reference--correction formulation of interaction anticipation, with a frozen geometry reference and zero-initialised residuals, under which the contribution of each cue is an exact, paired quantity and freezing can be examined as a single experimental factor.
\item A bidirectional cross-dataset measurement, with source-only model selection, date-level uncertainty and strong simple baselines, of how geometry, body pose and head orientation transfer between two robots and sites, including a check against a stronger, source-selected geometry reference.
\item An analysis of which cues can be expected to carry over to a new platform, which reverse, and how far rankings and thresholds transfer.
\end{itemize}

\section{Related work}\label{sec:related}

\subsection{Anticipating interaction with a robot}
Engagement, the process by which interaction partners establish and maintain a connection \citep{sidner2005engagement,oertel2020engagement}, has been estimated from position and orientation relative to the robot \citep{michalowski2006spatial}, from directed and mutual gaze \citep{rich2010recognizing}, from proxemic, body and face features \citep{vaufreydaz2016starting}, with classifiers for a multi-party bartender robot \citep{foster2017engagement}, and with metrics built from head and body posture \citep{anzalone2015engagement}. Field studies in shopping malls turned trajectory-based estimates of who is willing to interact into approach behaviours \citep{satake2009approach,kato2015may}, proxemics has been modelled for socially aware navigation \citep{riosmartinez2015proxemics,mead2017proxemics}, and greeting strategies for first encounters with mobile robots have been compared in the wild \citep{avelino2025greeting}. These works design or evaluate the robot's response; the present study concerns the perception estimate that such behaviours consume.

Recent anticipation models are data-driven. \citet{abbate2024self} let a service robot label its own encounters by whether an interaction eventually happened and trained classifiers on pose and motion features; adding gaze features to such a self-supervised classifier improved its ranking and the distance at which it was accurate \citep{arreghini2024predicting}, and a deployment in two crowded public locations compared proactive behaviours triggered by distance and by behavioural cues \citep{arreghini2024service}. PAR-D \citep{thompson2024pard} collects approach data around a public robot for intent prediction; \citet{bian2025interact} forecast intent to interact jointly with attitude and social action from whole-body skeletons; and MINT-RVAE \citep{mohsen2025mint} predicts frame-level interaction intent from pose and emotion cues. HUI360 \citep{lorenzolouis2026hui360} provides 360\textdegree{} egocentric recordings from a mobile robot with tracked persons, poses and masks, defines interaction as physical contact, reports baselines and feature ablations, and evaluates across datasets against SSUP-A. Forecasting human motion for robot behaviour generation uses the same ingredients, body pose and position over time \citep{vallsmascaro2025behavior}; trajectory prediction benefits from box and pose cues and can be made robust to missing ones \citep{saadatnejad2024social}, and datasets recorded from a social robot annotate the actions, groups and poses of the surrounding people \citep{ehsanpour2022jrdb,vendrow2023jrdb}. We adopt the task, data and bidirectional evaluation of \citet{lorenzolouis2026hui360}. Our model differs from these predictors in retaining a separately trained geometry prediction as an unchanged term of the final score, and our question differs in asking which cue carries over, not which model scores highest.

\subsection{Head orientation as an attention cue}
Face and head orientation are standard evidence of attention in engagement models \citep{michalowski2006spatial,sidner2005engagement,rich2010recognizing,anzalone2015engagement}, and gaze is reviewed as an early cue of intention in HRI \citep{belardinelli2024gaze}; mutual gaze has been detected at distances of up to 5\,m for this purpose \citep{arreghini2024long}. Most of these systems observe a person who is already near the robot or in conversation with it, often with a frontal camera. In the anticipation setting the person is further away, seen by a panoramic camera mounted on a moving platform, and must be judged before any exchange starts. We measure head orientation relative to the camera from dense face keypoints \citep{khirodkar2024sapiens}; we do not estimate gaze direction, and a person facing the camera is not necessarily looking at the robot.

\subsection{Combining cues on a fixed base}
In jointly trained multi-branch classifiers, branches overfit and generalise at different rates \citep{wang2020makes}, and a dominant modality suppresses the optimisation of the others \citep{peng2022balanced,zhang2024multimodal}. In work on dataset bias, \citet{he2019unlearn} fit a biased classifier, freeze it, and train an additional classifier through the loss of the summed scores, which is, in its training mechanics, the nearest neighbour of FRPR. Their goal is to discard the biased branch at test time, whereas our reference is a legitimate cue that stays in the deployed model. Learning a correction to a fixed predictor is also the principle of gradient boosting \citep{friedman2001greedy}, which has been used to train modalities in alternation \citep{hua2024reconboost}. None of these components is new; we use their combination to measure what a cue adds to a geometry-based prediction and whether that addition survives a change of robot and site. Our setting is that of domain generalisation, in which no target data are available during training \citep{zhou2023domain}; because cross-domain claims are sensitive to model selection \citep{gulrajani2021search}, all selection uses source data only. Skeleton-based recognisers are known to lose accuracy across datasets \citep{wang2025dual}, which motivates measuring the transfer of the pose cue separately.

\section{Fixed-reference pose residual model}\label{sec:method}

\subsection{Task and overview}\label{sec:task}
A sample is one tracked person observed for $T=32$ consecutive frames at 15\,fps, about 2.1\,s. For every frame $t$ the datasets provide a bounding box, the area of the segmentation mask and 17 body keypoints with confidences, in image-normalised coordinates of the equirectangular panorama. The label $y\in\{0,1\}$ states whether the person makes physical contact with the robot after the observation. Following \citet{lorenzolouis2026hui360}, an interaction starts when the person's mask intersects the interaction zone of the platform; a positive window ends a fixed lead time before that onset (16 frames, 1.07\,s, in the main experiments), and a negative window ends the same number of frames before the moment at which a non-interacting person appears largest. Windows are thus anchored on events, and the task is the binary classification of event-anchored prefixes, not continuous detection over a video stream. The negative anchor is determined from the complete track, including frames after the window; this rule of the official protocol is retained for comparability.

\begin{figure*}[t]
  \centering
  \includegraphics[width=\textwidth]{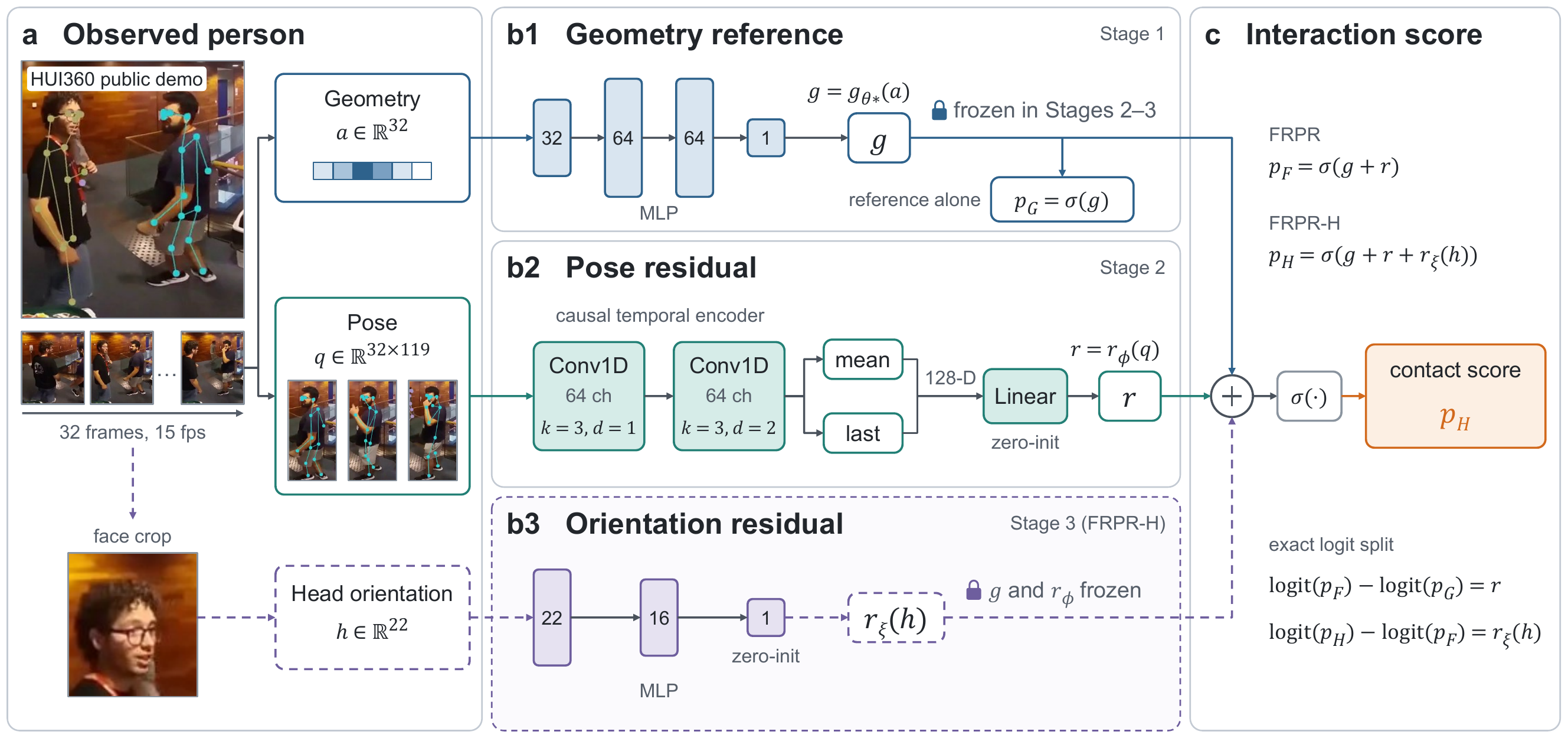}
  \caption{Fixed-reference pose residual (FRPR) and its head-orientation extension (FRPR-H). (a) A 32-frame track at 15\,fps yields a geometry summary $a$, a box-normalised pose sequence $q$ and, for FRPR-H only, a head-orientation summary $h$ from dense face keypoints; all inputs are standardised with source-training statistics. (b1) Stage~1 trains the geometry multilayer perceptron (MLP), whose logit $g=\gstar(a)$ gives the stand-alone reference score $p_G=\sigma(g)$; it remains frozen in Stages~2 and~3. (b2) Stage~2 trains a causal temporal encoder with a zero-initialised linear head, producing the pose residual $r=\rres(q)$. (b3) Stage~3, used by FRPR-H only, trains a small MLP with a zero-initialised output on $h$ while $g$ and $r$ stay frozen. (c) The logits are summed before the sigmoid, so each score splits exactly into the terms shown. Frames and crops are taken from the public HUI360 demonstration material; the coloured feature cells are schematic}
  \label{fig:overview}
\end{figure*}

From each window we compute a geometry summary $a\in\mathbb{R}^{32}$ and a pose sequence $q\in\mathbb{R}^{T\times119}$. A model maps $(a,q)$ to a logit $z$ and a score $p=\sigma(z)$. Training uses a labelled source domain, divided by recording date into $\Dtr$ and $\Dval$; the target domain $\Dtgt$ is used for evaluation only. FRPR (Fig.~\ref{fig:overview}) computes
\begin{equation}\label{eq:frpr}
z=\gstar(a)+\rres(q),\qquad p=\sigma(z),
\end{equation}
where $\gref$ is the geometry reference, $\theta^{*}$ its parameters after the first training stage, and $\rres$ the pose residual, trained in the second stage with $\theta^{*}$ held fixed. The residual branch receives the pose sequence only; it sees neither $a$ nor the reference logit.

\subsection{Input representation}\label{sec:inputs}
The event-anchored windows and the two representations are illustrated in Online Resource~1 (Fig.~S3); inputs are computed from observed frames only.

\textit{Geometry summary.} Let $(c^{x}_{t},c^{y}_{t})$ be the box centre, $w_{t}$ and $h_{t}$ its width and height, and $s_{t}$ the mask area as a fraction of the image. Boxes that cross the vertical seam of the panorama are unwrapped. The horizontal centre is an azimuth, $\psi_{t}=2\pi c^{x}_{t}$, encoded by its sine and cosine; its angular velocity $\omega_{t}$ is the finite difference of the unwrapped azimuth over the frame timestamps. Eight channels are formed per frame,
\begin{equation}\label{eq:geomchannels}
\begin{aligned}\bigl(&\sin\psi_{t},\cos\psi_{t},c^{y}_{t},\log w_{t},\log h_{t},\\&\log (w_{t}h_{t}),\log s_{t},\omega_{t}\bigr),\end{aligned}
\end{equation}
and each is reduced over the window to its last value, mean, standard deviation and least-squares slope against time, giving $a\in\mathbb{R}^{32}$. The summary describes where the person is, how large they appear and how both evolve; it contains apparent motion but no articulation.

\textit{Pose sequence.} A keypoint $j$ at frame $t$ is valid, $m_{t,j}=1$, if its coordinates are finite and inside the image and its confidence $\kappa_{t,j}$ is at least 0.05. Valid keypoints are expressed relative to the box, $\rho_{t,j}=\bigl((k^{x}_{t,j}-c^{x}_{t})/w_{t},(k^{y}_{t,j}-c^{y}_{t})/h_{t}\bigr)$, with the horizontal offset wrapped across the seam; invalid ones are set to zero. With the velocity $\dot\rho_{t,j}$ and its validity flag $\tilde m_{t,j}$, each keypoint contributes $(\rho_{t,j},\kappa_{t,j},\dot\rho_{t,j},m_{t,j},\tilde m_{t,j})$, so that $q_{t}\in\mathbb{R}^{119}$. All inputs are standardised with statistics of $\Dtr$, and missing keypoints remain exactly zero. The split into geometry and pose is operational: confidence and visibility still vary with distance and viewpoint.

\subsection{Geometry reference and pose residual}\label{sec:reference}
The reference $\gref$ is an MLP with two hidden layers of width 64, ReLU activations and dropout of 0.1 after the first hidden layer (6{,}337 parameters). It is trained with the binary cross-entropy $\bce(z,y)=-y\log\sigma(z)-(1-y)\log(1-\sigma(z))$,
\begin{equation}\label{eq:stage1}
\theta^{*}\approx\arg\min_{\theta}\sum_{(a,y)\in\Dtr}\bce\bigl(\gref(a),y\bigr),
\end{equation}
and $\theta^{*}$ is the checkpoint with the highest AP on $\Dval$. The reference is a complete predictor: its score $p_{G}=\sigma(\gstar(a))$ is the geometry-only model of our experiments.

The residual branch is a causal temporal convolutional network (TCN) \citep{bai2018tcn}: two one-dimensional convolutions with 64 channels, kernel size 3 and dilations 1 and 2, left padding only, so that the feature at frame $t$ depends on frames $t-6,\dots,t$. The feature sequence is summarised by its temporal mean and its last frame, and a linear head maps the 128-dimensional vector to the scalar $\rres(q)$ (35{,}393 parameters). The weight and bias of the head are initialised to zero, hence $\rres\equiv0$ before the first update and Eq.~\eqref{eq:frpr} starts the second stage as the reference itself.

\subsection{Two-stage training}\label{sec:training}
Stage~2 minimises the cross-entropy of the summed logit over $\phi$ alone,
\begin{equation}\label{eq:stage2}
\phi^{*}\approx\arg\min_{\phi}\sum_{(a,q,y)\in\Dtr}\bce\bigl(\gstar(a)+\rres(q),y\bigr).
\end{equation}
The residual is whatever logit offset reduces the classification loss given the reference; samples that the reference already classifies confidently contribute little gradient. The reference receives no gradient and is kept in evaluation mode, so $\gstar(a)$ is deterministic. Both stages use AdamW \citep{loshchilov2019adamw} and early stopping on the AP of $\Dval$; the decision threshold $\delta$ maximises $F_{1}$ on $\Dval$. Online Resource~1 (Fig.~S4) contrasts this procedure with a matched joint model that starts from the same checkpoint and differs only in updating the geometry parameters. The full algorithm, optimiser settings and training cost are given in Online Resource~1 (Sect.~S1).

\subsection{What the fixed reference provides}\label{sec:properties}
\textit{An exact split.} With $p_{G}=\sigma(\gstar(a))$, Eq.~\eqref{eq:frpr} is equivalent to
\begin{equation}\label{eq:identity}
\operatorname{logit}(p)-\operatorname{logit}(p_{G})=\rres(q).
\end{equation}
A positive residual multiplies the odds of interaction given by the reference by $e^{\rres(q)}$. The identity is algebraic and carries no causal meaning; its use is as a measurement. Both terms are stored for every prediction, and each can be scored as a ranking function on any subset of data (Section~\ref{sec:components}).

\textit{A nested pair of models.} FRPR contains its own reference, with identical weights, as the special case $\phi=\phi_{0}$. The difference in a metric between FRPR and the reference is therefore paired for every training seed and attributable to the pose branch, free of the seed-to-seed variation of separately trained models. The reference also remains available as a fallback when pose estimation fails.

\textit{Anchoring.} A jointly trained additive model $\gref(a)+\rres(q)$ has the same algebraic form, but the loss depends on $\theta$ and $\phi$ through the sum only, so logit mass can move between the branches without changing predictions, and $\gref$ need not resemble any model trained to predict from geometry. Freezing removes this freedom. It does not make the summed score at least as good as the reference on new data, and the unit weight of the sum fixes the relative scale of the two terms on the source domain.

\subsection{Controlled variants}\label{sec:variants}
Seven further neural models share the two building blocks (listed with their parameter counts in Online Resource~1, Table~S15). The geometry reference and the pose network alone measure each input. Three jointly trained additive models have exactly the architecture of FRPR. \emph{Joint, warm start} loads the same $\theta^{*}$ and the same initial $\phi$ as FRPR for a given seed and differs from it only in letting $\theta$ receive gradients; with shared optimiser settings, freezing is the only factor that separates them. \emph{Joint, warm + dropout} additionally re-activates geometry dropout, the usual way of fine-tuning, and \emph{joint, from scratch} trains both branches from random initialisation. Two nonlinear fusion models concatenate the penultimate geometry and pose features and apply a hidden layer of width 16; one has a pose network of width 64, the other of width 60 so that its parameter count matches FRPR within 0.4\%.

\subsection{Extension: a head-orientation residual}\label{sec:orientation}
A further cue $h$ can be attached as a third stage whose output is added to the logit of a frozen FRPR model,
\begin{equation}\label{eq:frprh}
z_{H}=\gstar(a)+r_{\phi^{*}}(q)+\rori(h),\qquad p_{H}=\sigma(z_{H}),
\end{equation}
where $\rori$ is trained with $\theta^{*}$ and $\phi^{*}$ held fixed. We call this model FRPR-H. The cue is the orientation of the head relative to the camera, which $q$ represents only implicitly through the nose, eye and ear keypoints. It was selected among several candidate cues by a screen on source data only (Section~\ref{sec:screen}).

\textit{Head-orientation summary.} Both datasets also provide 308 keypoints per person and frame in the layout of Sapiens \citep{khirodkar2024sapiens}, including dense face landmarks. Six channels are computed per frame from confident landmarks, relative to the box: the eye order $e_{t}=x^{\mathrm{Leye}}_{t}-x^{\mathrm{Reye}}_{t}$, positive when the face is turned towards the camera; a yaw asymmetry $\upsilon_{t}$ of the nose tip relative to lateral face landmarks, and $|\upsilon_{t}|$; a pitch ratio; the left--right ear-confidence difference; and the mean face-landmark confidence. Each is summarised by mean, last valid value and slope, and four values are appended (yaw variability, the fractions of frames with valid eye order and yaw, and the fraction of frontal frames), giving $h\in\mathbb{R}^{22}$ (Online Resource~1, Sect.~S7). The summary describes orientation relative to the camera, not gaze.

\textit{Orientation residual.} $\rori$ is an MLP with 16 hidden ReLU units, dropout of 0.1 and a zero-initialised output (\ORParTrFH{} trainable parameters), trained with the rules of Section~\ref{sec:training}. At the start of stage~3 FRPR-H equals the FRPR model of the same seed; the reference, FRPR and FRPR-H form a nested chain, and each prediction splits exactly into $g$, $r_{q}=r_{\phi^{*}}(q)$ and $r_{h}=\rori(h)$. Three comparators share the orientation encoder (Online Resource~1, Table~S15): FRPR-HK adds a hand summary $k$ to $h$; FRPR-QH learns the pose and orientation terms together on the frozen reference, testing whether fixing the pose term first matters; and Fusion-H extends the fusion model by $h$ and trains all parameters jointly.

\section{Experimental setup}\label{sec:setup}

\subsection{Datasets and sample pool}\label{sec:data}
HUI360 was recorded by a custom service robot with a 360\textdegree{} camera in several public environments over three months; SSUP-HRI was recorded by two mobile trash-barrel robots with 360\textdegree{} cameras on public squares in New York in two field sessions one year apart \citep{bu2024ssup}, and SSUP-A is its annotation released with HUI360 \citep{lorenzolouis2026hui360}. Both provide, per tracked person, boxes, masks, ViTPose keypoints \citep{xu2022vitpose} and interaction onsets defined by contact with the platform. The robots, camera heights, sites and crowd densities differ, and so does the prevalence of interaction: 3.0\% of the 4875 SSUP-A target windows and 16.7\% of the 407 HUI360 target windows are interactions (all partitions in Online Resource~1, Table~S16). We export one 32-frame window per track with the official sampling rule, so that all models see identical samples. The official training pool of the source dataset is split by recording date into training and validation sets, and the official test pool of the other dataset is the target. The HUI360 test pool comes from held-out environments, the SSUP-A test pool from the second field session. Two directions are studied: \fwd{} and \rev.

Two properties of the official pipeline affect interpretation. First, the official loader selects negative windows by comparing two frame positions expressed in different coordinates, so negative tracks with a missing frame before the anchor are discarded; these are mostly long tracks (median length \PFLenAddHUISrc{} frames on the HUI360 source against \PFLenNegHUISrc{} for retained negatives and \PFLenPosHUISrc{} for positives). All main experiments use the official pool for comparability with the benchmark. As a sensitivity analysis we rebuilt the pool with one frame-coordinate rule for both classes, which adds \PFAddedNeg{} negative windows and leaves all \PFPositives{} positives unchanged, and retrained the main experiment on it (Online Resource~1, Sect.~S6). Second, the geometry summary uses the absolute azimuth in the panorama. The official loader can instead centre each frame on the robot's interaction zone; on HUI360 this shift is zero, but on SSUP-A the zone centre moves across the panorama, so there the azimuth channels describe the direction in the image rather than relative to the robot (details in Online Resource~1, Sect.~S6).

\subsection{Compared methods}\label{sec:baselines}
Besides the neural models above, we include a random forest \citep{breiman2001random} with 200 class-balanced trees and a histogram-based gradient boosting classifier (HistGB) with 150 iterations \citep{pedregosa2011scikit}. Trees receive summary statistics $\bar q$ of the standardised pose sequence (last frame, temporal mean and standard deviation; 357 values), and each ensemble is trained on $a$, on $\bar q$ and on both, to separate the learner from the input. The simplest competitor has no learned parameters: it ranks target windows by the logarithm of the box area at the last observed frame, with the sign chosen on source validation data. The published HUI360 baselines use a different sampling and normalisation protocol and are not compared numerically.

\subsection{Model selection and statistical analysis}\label{sec:selection}
Every choice that could favour a method is made on source data (Online Resource~1, Fig.~S1). Hyperparameters are selected by three-fold cross-validation inside $\Dtr$ with folds formed by recording date, two seeds per fold, and mean AP as criterion. Each neural model searches learning rates $\{3\times10^{-4},10^{-3},3\times10^{-3}\}$ and weight decays $\{5\times10^{-5},5\times10^{-4}\}$; FRPR and the warm-started models reuse the reference configuration and search the grid for stage~2. Tree ensembles search six configurations each, are fitted on $\Dtr$ and use $\Dval$ only for the threshold (refitting on $\Dtr\cup\Dval$: Online Resource~1, Sect.~S4). Five models are trained with seeds 11 to 55 (HistGB once). Among the joint additive models and among the fusion models, the one with the highest source cross-validation AP is the ``best joint'' and ``best fusion'' competitor of that direction. Target predictions were computed once, after all configurations and checkpoints had been fixed.

The primary metric is AP, which suits rare positives \citep{saito2015precision}; because prevalence differs between the targets, AP is not compared across directions. We also report the area under the ROC curve (AUROC) and, at the source-selected threshold, recall and false-positive rate (FPR); for a robot these operating-point metrics correspond to missed interactions and false alarms. Methods are compared through paired differences $\dAP$, the mean over seeds of $\mathrm{AP}_{A}-\mathrm{AP}_{B}$ on the same target samples. Windows from one recording day are not independent, so each of 10{,}000 bootstrap replicates resamples target dates with replacement, keeps all windows of a drawn date, resamples the five seeds and recomputes $\dAP$ \citep{efron1994bootstrap,field2007bootstrapping}; we report percentile 95\% intervals and leave-one-date-out differences. With five and nine target dates these intervals are approximate, and an interval containing zero is read as absence of evidence for a difference, not as equivalence.

Four primary comparisons were fixed in an internal, hash-locked protocol before any model of this study was scored on target data: FRPR against the best joint and the best fusion model, in each direction; for these we also give 98.75\% (Bonferroni) intervals. All other differences, including FRPR against its reference, are exploratory with unadjusted 95\% intervals; the box-area rule and the input-specific tree ensembles were added after the first results had been seen. Both target sets had been evaluated in earlier stages of this project, so the study is not a blind test, although no target information entered training, selection or thresholding.

\subsection{Screening and locked evaluation of the orientation extension}\label{sec:screen}
Candidate cues were screened on source data only, under pass criteria fixed before any candidate model was fitted. Each candidate block was appended to HistGB on $a$ and $\bar q$: the head-orientation summary $h$, a hand summary $k$, both together, the Sapiens body keypoints in place of ViTPose, a mask-shape summary, and the azimuth relative to the interaction zone; confidence-only summaries and $h,k$ permuted within dates served as controls. A block passed if, on held-out source dates of both source domains, it raised AP by at least 0.015, improved more than half of the qualifying dates and, for blocks containing $h$ or $k$, gained at least twice as much as each control (details in Online Resource~1, Sect.~S7). FRPR-H, its comparators and all settings were then fixed in a second protocol before head summaries were extracted for any target window, and all checkpoints, features and code were hashed before a single target scoring. The protocol's decision rule required FRPR-H to exceed FRPR in mean $\dAP$ in both directions with at least four of five seeds positive, and a 95\% interval above zero in at least one direction and below zero in none. Tree ensembles with $h$ appended were locked separately and scored after the results of FRPR-H had been obtained.

\section{Results}\label{sec:results}

\subsection{Main comparison}\label{sec:main}
Table~\ref{tab:main} gives the target-domain results; the AP of every training seed is plotted in Online Resource~1 (Fig.~S5). Three observations organise this section.

First, the value of the pose correction depends on the direction. In \rev{} FRPR reaches an AP of \APFRev{} against \APGRev{} for its reference, and all five seeds improve. In \fwd{} the two mean APs agree to three decimals (\APFFwd{} and \APGFwd). Models that see only pose are close to chance in \fwd{} (AP \APQFwd{} to \APHGBposeFwd{} at a prevalence of 0.030) and as good as or better than geometry-only models in \rev.

Second, how the inputs are combined matters less than which inputs are available: the six neural models that use both inputs lie within a few hundredths of AP of each other, of the order of the seed-to-seed standard deviation, and FRPR is not the best of them in either direction.

Third, simple models are strong. Ranking by last-frame box area gives the highest AP in \fwd{} (\APAreaFwd). Tree ensembles are the best models in \rev, where HistGB with both inputs reaches \APHGBallRev{} and \TPHGBaqHAcvRev{} with the head-orientation summary appended. Within the geometry input the random forest outperforms the reference network in \rev{} (\APRFgeoRev{} against \APGRev), so the reference of the main study is not the strongest geometry predictor available; Section~\ref{sec:strongref} examines what this means for the pose gain.

The operating-point columns add a caution for every learned model. At the threshold chosen on source validation data, the neural models that use geometry detect at most 4\% of the target interactions in \fwd{} and at most 17\% in \rev: their scores shrink under the domain change, and a source-appropriate threshold becomes far too strict. Rankings, which AP and AUROC measure, transfer much better than thresholds.

\begin{table*}[t]
\caption{Target-domain results under source-only model selection. AP and AUROC are means over five training seeds ($\pm$ standard deviation of AP); recall and false-positive rate (FPR) are taken at the threshold that maximises $F_{1}$ on source validation data. $\bar q$: summary statistics of the pose sequence; $h$, $k$: head-orientation and hand summaries. The last block contains the orientation extension and HistGB with $h$ appended, each scored once after its own lock. Bold and underline: highest and second-highest AP per direction; shading: FRPR and FRPR-H. AP is not comparable between directions because prevalence differs (0.030 and 0.167)}\label{tab:main}
\footnotesize\setlength{\tabcolsep}{2.5pt}\renewcommand{\scriptsize}{\fontsize{6pt}{7pt}\selectfont}
\begin{tabular*}{\textwidth}{@{\extracolsep{\fill}}ll cccc cccc@{}}
\toprule
 & & \multicolumn{4}{c}{\fwd} & \multicolumn{4}{c}{\rev} \\
\cmidrule(lr){3-6}\cmidrule(l){7-10}
Method & Input & AP $\uparrow$ & AUROC $\uparrow$ & Recall $\uparrow$ & FPR $\downarrow$ & AP $\uparrow$ & AUROC $\uparrow$ & Recall $\uparrow$ & FPR $\downarrow$ \\
\midrule
Last-frame box area\tnote{a} & $a$ & \textbf{0.178} & 0.858 & 0.61 & 0.133 & 0.341 & 0.732 & 0.01 & 0.012 \\
\addlinespace[2pt]
Geometry reference & $a$ & 0.129\,{\scriptsize$\pm$0.014} & 0.841 & 0.03 & 0.004 & 0.277\,{\scriptsize$\pm$0.022} & 0.733 & 0.00 & 0.006 \\
Random forest & $a$ & 0.146\,{\scriptsize$\pm$0.006} & 0.860 & 0.19 & 0.028 & 0.369\,{\scriptsize$\pm$0.011} & 0.738 & 0.15 & 0.031 \\
HistGB\tnote{b} & $a$ & 0.161 & 0.841 & 0.19 & 0.022 & 0.311 & 0.688 & 0.19 & 0.047 \\
\addlinespace[2pt]
Pose TCN & $q$ & 0.048\,{\scriptsize$\pm$0.002} & 0.652 & 0.21 & 0.119 & 0.334\,{\scriptsize$\pm$0.009} & 0.757 & 0.21 & 0.080 \\
Random forest & $\bar q$ & 0.050\,{\scriptsize$\pm$0.001} & 0.651 & 0.09 & 0.038 & 0.348\,{\scriptsize$\pm$0.005} & 0.749 & 0.18 & 0.046 \\
HistGB\tnote{b} & $\bar q$ & 0.055 & 0.685 & 0.36 & 0.180 & 0.416 & 0.794 & 0.15 & 0.015 \\
\addlinespace[2pt]
\textbf{FRPR} (fixed reference) & $a,q$ & \cellcolor{ourtint}0.129\,{\scriptsize$\pm$0.011} & 0.836 & 0.04 & 0.006 & \cellcolor{ourtint}0.321\,{\scriptsize$\pm$0.035} & 0.761 & 0.01 & 0.006 \\
Joint, warm start & $a,q$ & 0.142\,{\scriptsize$\pm$0.017} & 0.858 & 0.02 & 0.004 & 0.326\,{\scriptsize$\pm$0.014} & 0.767 & 0.02 & 0.009 \\
Joint, warm + dropout & $a,q$ & 0.137\,{\scriptsize$\pm$0.018} & 0.854 & 0.02 & 0.003 & 0.336\,{\scriptsize$\pm$0.024} & 0.774 & 0.01 & 0.005 \\
Joint, from scratch & $a,q$ & 0.119\,{\scriptsize$\pm$0.024} & 0.816 & 0.02 & 0.004 & 0.357\,{\scriptsize$\pm$0.047} & 0.770 & 0.17 & 0.040 \\
Fusion, width 64 & $a,q$ & 0.149\,{\scriptsize$\pm$0.015} & 0.859 & 0.02 & 0.003 & 0.325\,{\scriptsize$\pm$0.029} & 0.757 & 0.04 & 0.014 \\
Fusion, matched size & $a,q$ & 0.145\,{\scriptsize$\pm$0.014} & 0.864 & 0.01 & 0.002 & 0.326\,{\scriptsize$\pm$0.015} & 0.751 & 0.06 & 0.016 \\
Random forest & $a,\bar q$ & 0.139\,{\scriptsize$\pm$0.007} & 0.855 & 0.02 & 0.002 & 0.399\,{\scriptsize$\pm$0.006} & 0.755 & 0.25 & 0.061 \\
HistGB\tnote{b} & $a,\bar q$ & 0.129 & 0.849 & 0.01 & 0.003 & \underline{0.427} & 0.789 & 0.32 & 0.068 \\
\addlinespace[2pt]
\textbf{FRPR-H} & $a,q,h$ & \cellcolor{ourtint}0.140\,{\scriptsize$\pm$0.017} & 0.844 & 0.04 & 0.005 & \cellcolor{ourtint}0.328\,{\scriptsize$\pm$0.028} & 0.767 & 0.01 & 0.006 \\
FRPR-HK & $a,q,h,k$ & 0.131\,{\scriptsize$\pm$0.011} & 0.838 & 0.04 & 0.006 & 0.350\,{\scriptsize$\pm$0.042} & 0.780 & 0.03 & 0.009 \\
FRPR-QH & $a,q,h$ & 0.151\,{\scriptsize$\pm$0.016} & 0.869 & 0.03 & 0.004 & 0.315\,{\scriptsize$\pm$0.029} & 0.760 & 0.00 & 0.003 \\
Fusion-H & $a,q,h$ & \underline{0.173}\,{\scriptsize$\pm$0.017} & 0.887 & 0.01 & 0.002 & 0.340\,{\scriptsize$\pm$0.019} & 0.779 & 0.02 & 0.009 \\
HistGB\tnote{b} & $a,\bar q,h$ & 0.154 & 0.860 & 0.30 & 0.037 & \textbf{0.525} & 0.809 & 0.35 & 0.068 \\
\bottomrule
\end{tabular*}
\tabnote{a}{Uses one entry of $a$, the logarithm of the box area at the last frame; no learned parameters, sign and threshold chosen on source validation data.}
\tabnote{b}{Deterministic given the data; trained once.}
\end{table*}

\subsection{Value of the pose correction over its reference}\label{sec:value}
\begin{figure*}[t]
  \centering
  \includegraphics[width=\textwidth]{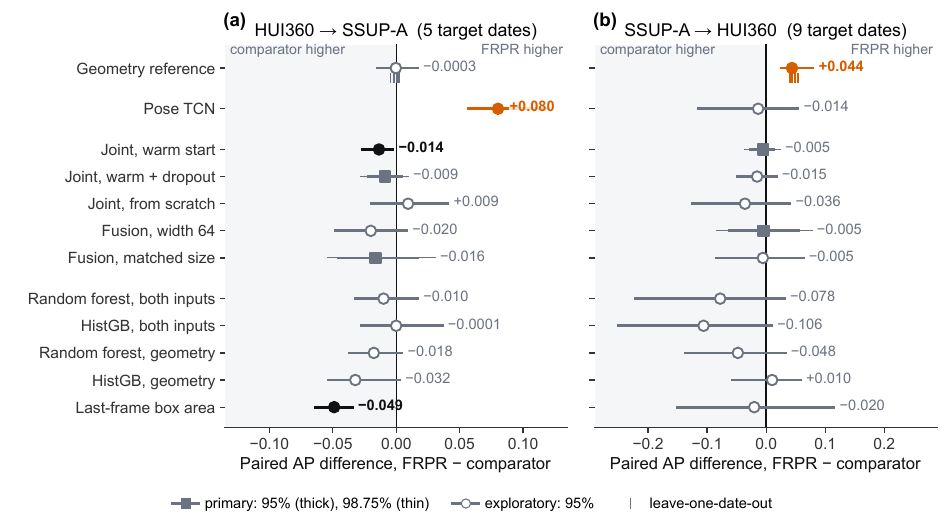}
  \caption{Paired AP differences between FRPR and each comparator on the target domain. Thick bars are 95\% percentile intervals from a bootstrap over recording dates and training seeds. Squares mark the four primary comparisons; for these only, thin bars give the 98.75\% intervals of the Bonferroni family. Circles mark exploratory comparisons with unadjusted 95\% intervals; filled markers indicate intervals that exclude zero. Ticks under the first row are the differences obtained when one target date is removed. Horizontal scales differ between panels; all values are listed in Online Resource~1 (Table~S3)}
  \label{fig:forest}
\end{figure*}

In \rev{} the paired difference between FRPR and its reference is \DFGRev{} AP with a 95\% interval of \CIFGRev{} (Fig.~\ref{fig:forest}b). Removing any one of the nine target dates leaves a difference between \LodoMinRev{} and \LodoMaxRev, and the seed-averaged difference is positive for each of five alternative groups of source validation dates (Online Resource~1, Fig.~S6b). In \fwd{} the difference is \DFGFwd{} \CIFGFwd, with leave-one-date-out values between \LodoMinFwd{} and \LodoMaxFwd, so no gain is observed; the interval is compatible with changes of about 0.02 AP either way. The main difference is robust to the sample pool: on the rebuilt pool FRPR exceeds its reference by \PFDFGRev{} \PFCIFGRev{} in \rev{} (\PFSeedsFGRev{} seeds positive) and differs from it by \PFDFGFwd{} \PFCIFGFwd{} in \fwd. On held-out source dates, FRPR has the highest cross-validated AP of all models with inputs $a$ and $q$ in both source domains (Online Resource~1, Sect.~S2).

\subsection{Freezing against joint training and fusion}\label{sec:freezing}
The four primary comparisons ask whether freezing the reference pays in AP against the best jointly trained additive model and the best fusion model of each direction. We find no such advantage: all four point estimates are negative (\DFJFwd{} and \DFCFwd{} in \fwd, \DFJRev{} and \DFCRev{} in \rev) and all four intervals contain zero (Fig.~\ref{fig:forest}), also on the rebuilt pool. Among the exploratory comparisons, the warm-started joint model with its own selected settings is higher than FRPR in \fwd{} (\DFJwFwd, \CIFJwFwd). When FRPR and this model share one optimiser configuration, so that freezing is the only difference, the paired difference is \DFixFJwFwd{} \CIFixFJwFwd{} in \fwd{} and \DFixFJwRev{} \CIFixFJwRev{} in \rev. Protecting the geometry branch thus neither improved cross-domain ranking nor showed a cost under matched settings.

FRPR is lower than the box-area ranking in \fwd{} (\DFAreaFwd, \CIFAreaFwd); in \rev{} this difference is uninformative (\DFAreaRev, \CIFAreaRev). HistGB with both inputs is higher than FRPR in \rev{} (\DFHGBallRev, \CIFHGBallRev), with an interval excluding zero on the rebuilt pool and when refitted on all source data; in \fwd{} the ordering between tree ensembles and FRPR depends on the pool and on refitting (Online Resource~1, Sects.~S4 and S6).

\subsection{What the components reveal}\label{sec:components}
\textit{Anchoring.} Online Resource~1 (Table~S4) compares the four additive models term by term. For the warm-started joint models the mean squared change of the geometry logit on target samples reaches up to 139 squared logit units, while the ranking quality of that branch changes by at most 0.005 AP. Joint training thus rescales and shifts the geometry logit without improving it as a predictor, and for these models the split $z=g+r$ has no fixed reading. In FRPR the change is zero by construction.

\begin{figure*}[t]
  \centering
  \includegraphics[width=\textwidth]{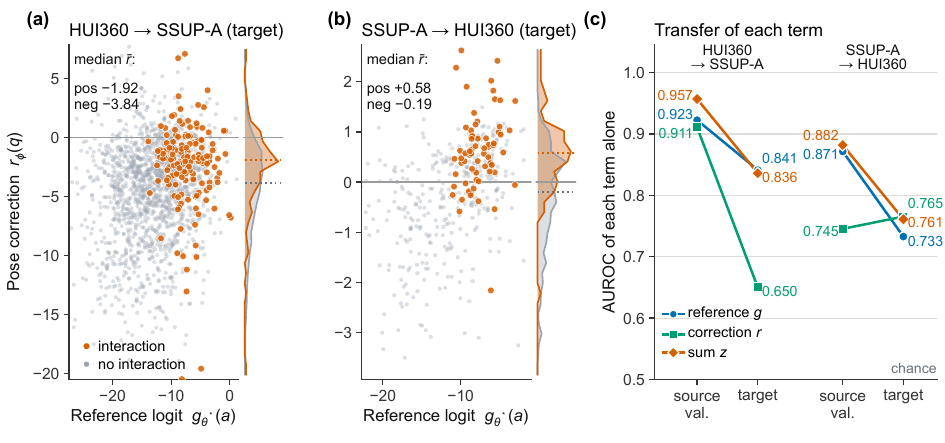}
  \caption{The two terms of FRPR on target data. (a, b) Reference logit against pose correction for every target sample (means over five seeds; 1500 randomly drawn non-interacting samples are shown in (a)), with the marginal distribution of the correction by class; dotted lines and printed values are class medians. (c) AUROC of each term scored alone, on source validation dates and on the target. In \fwd{} the reference keeps most of its ranking quality and the correction does not; in \rev{} the correction keeps its level while the reference drops and has the larger logit spread (note the different axis ranges in (b))}
  \label{fig:decomp}
\end{figure*}

\textit{Transfer of each term.} With the reference anchored, the two terms can be scored separately (Fig.~\ref{fig:decomp}c; AUROC is used because it is insensitive to the different prevalence of source and target). On HUI360 validation dates both terms are strong rankers (AUROC \AUCgValFwd{} and \AUCrValFwd). After transfer to SSUP-A the reference keeps most of its quality (\AUCgTgtFwd) while the pose correction falls to \AUCrTgtFwd, with heavily overlapping class distributions centred well below zero (Fig.~\ref{fig:decomp}a). In the other direction the roles are exchanged: on SSUP-A validation dates the correction is a modest ranker (\AUCrValRev) next to a strong reference (\AUCgValRev), and on HUI360 the reference drops to \AUCgTgtRev{} while the correction holds its level (\AUCrTgtRev). Geometry learned on HUI360 carries over to SSUP-A and pose does not; pose learned on SSUP-A carries over to HUI360 and geometry much less.

\textit{Relative scale.} In \rev{} the pose correction alone ranks the target better than the sum (AP \APrTgtRev{} against \APFRev). The reference logit has a standard deviation of \SDgRev{} units on the target and the correction one of \SDrRev{} (Fig.~\ref{fig:decomp}b), so the order of the summed scores largely follows the reference, the weaker term after transfer. This reading is exploratory: it was made on target data after the fact, and no combination weight was tuned or tested.

\subsection{Sensitivity to lead time and to the content of the pose input}\label{sec:sensitivity}
\textit{Lead time.} The selected models were applied without retraining to windows ending 1.07 to 3.20\,s before the onset, on tracks available at all five lead times (Online Resource~1, Sect.~S5 and Fig.~S6a). In \rev{} the gain of FRPR over its reference is positive at every lead time, with intervals above zero at three of the five; in \fwd{} it stays close to zero throughout. With 27 common-track interactions in \rev, this shows only that the finding is not specific to one lead time.

\textit{Pose content.} Three controls, run at a fixed configuration under which FRPR exceeds its reference by \DFixFGRev{} \CIFixFGRev{} in \rev, test what the correction uses (Online Resource~1, Figs.~S1d and S6c). With pose sequences permuted among the samples of each date, intact FRPR exceeds the permuted models in \rev{} by \FPermMinRev{} to \FPermMaxRev, each interval above zero, and the permuted models differ from the reference by only \PermGMinRev{} to \PermGMaxRev. With keypoint coordinates and velocities zeroed, intact FRPR is ahead by \DFixConfRev{} and the confidence-only control does not differ from the reference (interval \CIFixConfGRev). The gain therefore requires the keypoint positions of the right person and is not obtained from detection quality alone. Removing only the explicit velocity channels leaves a model that still exceeds the reference (\DFixNoVelGRev, \CIFixNoVelGRev). In \fwd{} the controls stay close to the intact model.

\subsection{Pose residuals on a stronger geometry reference}\label{sec:strongref}
The reference network of the main study is weaker than the random forest on geometry in \rev{} (Section~\ref{sec:main}), so the pose gain there could in principle compensate for a weak reference rather than add transferable information. We therefore repeated the residual construction on a reference selected on source data among three families. For each direction and seed, the source training dates were split into three outer date folds; within the training dates of each fold, a nested date-fold cross-validation chose among an MLP on $a$, HistGB and a random forest, each with a small grid, by mean AP. The chosen model, refitted and temperature-scaled, provided out-of-fold reference logits for the held-out fold, and at deployment the three fold references were averaged (``selected reference''). A pose residual of the same additive form as FRPR, with a zero-initialised pose head, was trained on these out-of-fold logits and early-stopped on source validation AP. A HistGB reference with one fixed configuration, built in the same way, served as a second reference. Target AP was computed once after all source runs had been locked (protocol in Online Resource~1, Sect.~S8). Paired intervals resample target dates only (\SRBoot{} replicates, seeds averaged within a replicate) and are not directly comparable to the date-and-seed intervals above. The analysis is post hoc and not blind: its arms and comparisons were fixed after earlier target results of this project, including a first version of the analysis, had been inspected.

Online Resource~1 (Table~S17) gives the full results. The selected reference reaches an AP of \SRAPSelRefFwd{} in \fwd{} and \SRAPSelRefRev{} in \rev, above the reference of the main study (\APGFwd{} and \APGRev). The pose residual on this reference reaches \SRAPSelFFwd{} and \SRAPSelFRev. Its gain over its own reference is distinguishable from zero in \rev{} (\SRDFRefRev, \SRCIFRefRev, \SRSeedsFRefRev{} seeds positive) and not in \fwd{} (\SRDFRefFwd, \SRCIFRefFwd). The pose residual on the HistGB reference shows the same pattern (\SRDHGFHGRefRev{} \SRCIHGFHGRefRev{} in \rev, \SRDHGFHGRefFwd{} \SRCIHGFHGRefFwd{} in \fwd). Within this analysis, the direction dependence of the pose gain is thus not an artefact of the weak reference network. Against HistGB with fixed settings on $a$ and $\bar q$, the selected-reference model is distinguishably higher in \fwd{} (\SRDFHGBallFwd, \SRCIFHGBallFwd) and not distinguishable in \rev{} (\SRDFHGBallRev, \SRCIFHGBallRev). The selected family was not stable: among the \SRSelNFwd{} outer folds of each direction, the MLP was chosen \SRSelMLPFwd{} and \SRSelMLPRev{} times, HistGB \SRSelHistGBFwd{} and \SRSelHistGBRev{} times and the random forest \SRSelRFFwd{} and \SRSelRFRev{} times (Online Resource~1, Sect.~S8), and in \rev{} the fixed HistGB reference with its residual scored higher than the selected one (\SRDFHGFRev, \SRCIFHGFRev). At source-selected thresholds, the selected-reference model detects only \SRRecSelFFwd{} and \SRRecSelFRev{} of the target interactions on average.

\subsection{Head orientation as a further residual}\label{sec:orientation-results}
\textit{Source screen.} Appending the head-orientation summary $h$ raised the date-averaged source AP of HistGB by \SCDHFwd{} on HUI360, with all \SCDatesFiveFwd{} qualifying dates improved, and by \SCDHRev{} on SSUP-A (\SCDatesHRev{} dates). The permutation and confidence controls changed it by at most \SCDCRev, so the gain requires the orientation values of the right person. The hand summary and the combined block also passed; the Sapiens body keypoints, the mask shape and the zone-relative azimuth did not (Online Resource~1, Table~S10).

\textit{Target results.} FRPR-H exceeds FRPR by \ORDFHFFwd{} AP in \fwd{} (95\% interval \ORCIFHFFwd, \ORSeedsFHFFwd{} seeds positive) and by \ORDFHFRev{} in \rev{} (\ORCIFHFRev, \ORSeedsFHFRev{} seeds), and the decision rule of the protocol returns ``supported''; all orientation comparisons are listed in Online Resource~1 (Table~S11). The gains are small; at the 98.75\% level the lower bound in \rev{} is \ORFHFBLowRev. In \fwd, where the pose correction added nothing, this is the first residual term that improves on the fixed reference. Relative to other models, FRPR-H does not stand out. Fusion-H has higher point estimates in both directions, FRPR-QH is ahead in \fwd{} and behind in \rev{} with intervals that include zero, and the box-area rule remains above FRPR-H in \fwd{} (\ORDFHAreaFwd, \ORCIFHAreaFwd). Tree ensembles gain more from $h$: appending it to HistGB adds \TPDHGBhAcvFwd{} \TPCIHGBhAcvFwd{} in \fwd{} and \TPDHGBhAcvRev{} \TPCIHGBhAcvRev{} in \rev, and in \rev{} HistGB with $h$ lies clearly above FRPR-H (\ORDFHHGBaqHRev, \ORCIFHHGBaqHRev). Head orientation is thus informative for several learners; FRPR-H is the model in which its contribution can be read exactly, not the model that exploits it best.

\begin{figure*}[t]
  \centering
  \includegraphics[width=\textwidth]{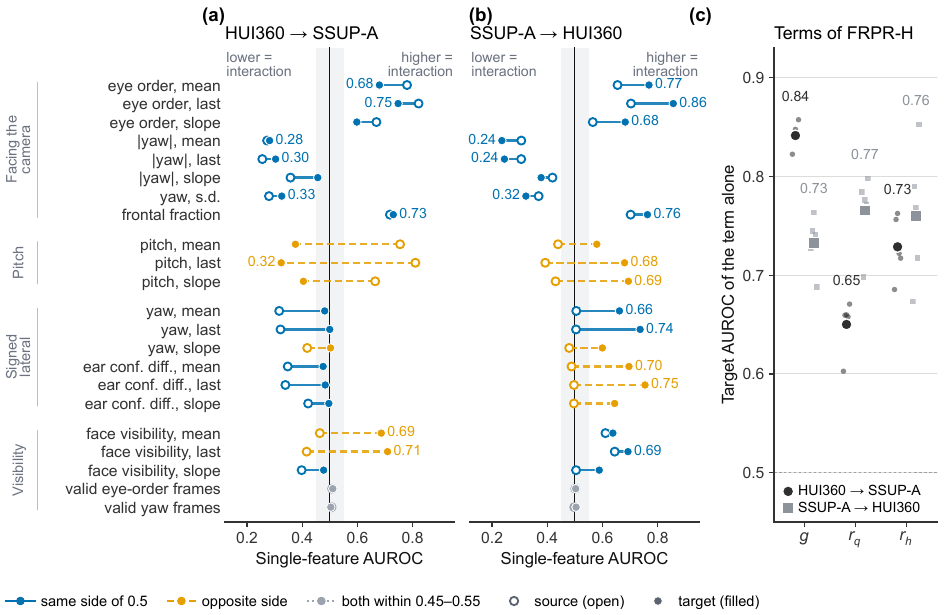}
  \caption{Transfer of the head-orientation cues and of the terms of FRPR-H. (a, b) AUROC of each entry of $h$ used alone as a score, on the source pool (open circles) and on the target set (filled circles); values below 0.5 mean that lower values indicate interaction. Colour and line style mark whether source and target lie on the same side of 0.5 (blue, solid), on opposite sides (orange, dashed), or both within 0.45--0.55 (grey, dotted); target values at least 0.15 from 0.5 are printed. Computed after the locked target scoring; post hoc and descriptive. (c) Target AUROC of the reference logit $g$, the pose term $r_{q}$ and the orientation term $r_{h}$ of FRPR-H, each scored alone; circles denote \fwd{} and squares \rev, small markers are seeds and large markers their mean}
  \label{fig:orientation}
\end{figure*}

\textit{The orientation term.} The new term is small (mean absolute value \ORrhAbsFwd{} and \ORrhAbsRev{} logit units, against \ORrqAbsFwd{} and \ORrqAbsRev{} for the pose term), but scored alone it ranks the target with an AUROC of \ORrhAUCFwd{} in \fwd, above the pose term (\ORrqAUCFwd), and \ORrhAUCRev{} in \rev, close to it (\ORrqAUCRev; Fig.~\ref{fig:orientation}c).

\textit{Which orientation cues transfer.} After the target scoring we computed the AUROC of each of the 22 entries of $h$ on its own, on the source pool and on the target set of each direction (Fig.~\ref{fig:orientation}a,b). This analysis was not planned in the protocol and was not used to change any model. The entries that describe whether a person faces the camera keep their discriminative direction in both transfers: interacting persons show a larger eye order, a smaller $|\upsilon|$, a steadier yaw and a larger fraction of frontal frames, with target AUROCs between \ORFaceAUCMin{} and \ORFaceAUCMax{} when oriented by the source sign. Head pitch reverses in both directions: its mean and last value separate the classes on the HUI360 source with AUROCs of \ORPitchSrcMinFwd{} to \ORPitchSrcMaxFwd{} and on the SSUP-A target with \ORPitchTgtMinFwd{} to \ORPitchTgtMaxFwd, that is, in the opposite direction. The signed lateral cues, yaw and ear-confidence difference, separate the classes only on HUI360 data, with opposite directions in its training and held-out environments, and are close to chance on SSUP-A. Of the 22 entries, \ORKeptFeatures{} keep their direction in both transfers.

\subsection{Individual corrections}\label{sec:qualitative}
\begin{figure*}[t]
  \centering
  \includegraphics[width=\textwidth]{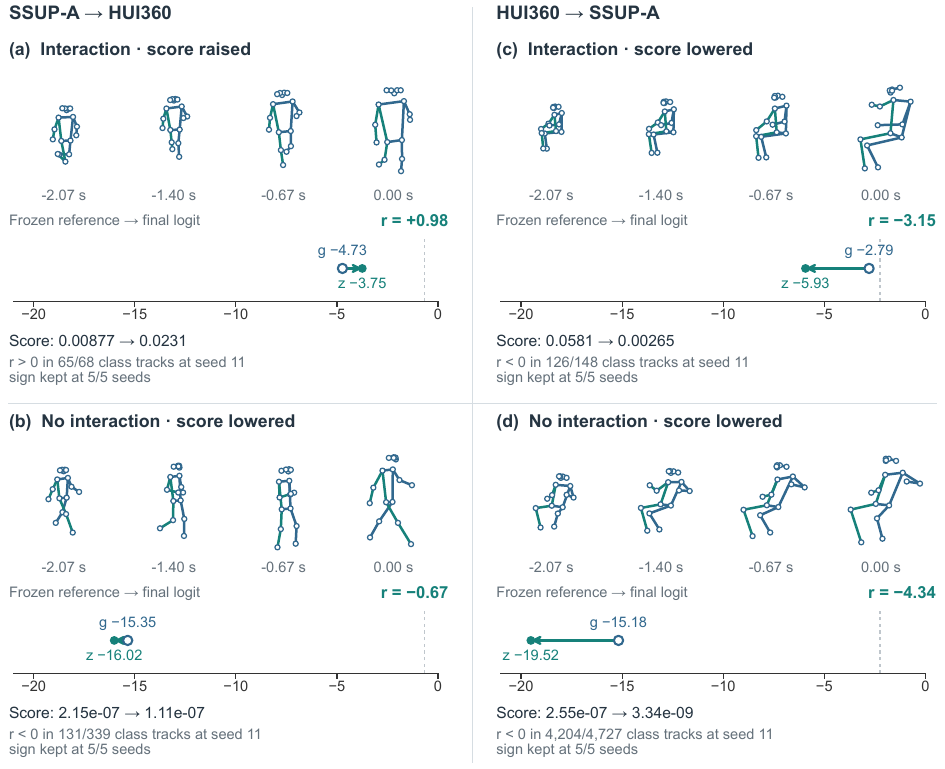}
  \caption{Observed pose sequences and their exact score decomposition under the frozen seed-11 models. Four frames are sampled from each 32-frame input; time zero is the observation cutoff, not the interaction onset. Skeletons use the recorded keypoints, are centred on the bounding box and share a spatial scale within each sequence. Open blue dots mark the frozen reference logit $g$, filled green dots the final logit $z=g+r$; dashed lines mark the source-selected FRPR threshold. Scores are sigmoid outputs, not calibrated probabilities. Counts give the number of target tracks of the indicated class with the displayed residual sign at seed 11; ``sign kept'' counts the seeds, of five, under which the displayed track has the same residual sign. Cases (a, b, d) move towards their labels and (c) away from it; neither positive case crosses its threshold}
  \label{fig:cases}
\end{figure*}

Figure~\ref{fig:cases} connects the logit decomposition to observed tracks. The cases come from the FRPR checkpoints of seed 11, fixed before any case was inspected. For each direction and label, the displayed stratum is the residual sign that most tracks of the class receive under the seed-averaged residual, and the track closest to the stratum median after a pose-visibility filter is shown; cases are thus selected by frequency, not by whether a score moves towards the label. In case~(a) an interacting HUI360 track receives a positive correction of \CaseAR, which raises its score but leaves it below the source-selected threshold, as under every seed. In \fwd{} both displayed corrections are negative, as is the bulk of the correction for both classes (Fig.~\ref{fig:decomp}a), so the interaction of case~(c) is down-scored. The residuals keep their sign under all five seeds but vary widely in size. Per-seed values and integrated-gradient attributions are given in Online Resource~1 (Sect.~S9, Fig.~S2); the attributions largely follow the displacement of the input from a static reference and are not read as evidence about particular body parts.

\section{Discussion}\label{sec:discussion}

\subsection{Which cues a social robot can carry to a new platform}
Geometry learned on HUI360 kept its ranking quality on SSUP-A, whereas geometry learned on SSUP-A lost much of it on HUI360; pose behaved the other way round. Neither cue can therefore be assumed to carry over in general. Camera placement and typical approach distances probably differ between the custom service robot of HUI360 and the trash-barrel robots of SSUP-A, and with them the apparent size at which a person is near; HUI360 also offers few training interactions (164) for a pose network of 35{,}393 parameters. The stronger-reference analysis removes one alternative explanation, a weak reference, since the asymmetry persisted with a source-selected reference. Testing the remaining explanations would require a third dataset or controlled subsampling.

Head orientation gave the most interpretable result. Whether a person faces the camera, and hence the robot that carries it, kept its discriminative direction across both robots and sites, in line with the use of body and face orientation as engagement cues \citep{michalowski2006spatial,sidner2005engagement,anzalone2015engagement}. Pitch reversed, which may reflect different camera heights relative to the faces of approaching people, and signed yaw separated the classes only on HUI360, where its direction differed between training and held-out environments, as one would expect if it depends on the side from which people pass the robot. These explanations are untested. For designers, the result suggests that cues describing attention towards the robot are candidates for transfer, whereas cues tied to camera geometry, such as pitch or the side of approach, need to be re-examined for each platform. A cue selected on source data can carry a site-specific component with the wrong sign; a screen over several source sites, or a restriction of $h$ to the facing descriptors, would test whether this limits FRPR-H.

\subsection{What anticipation errors mean for the robot}
Rankings and thresholds transferred differently: the models order people by their likelihood of contact above chance on a new platform, yet at thresholds chosen on source data the neural models that use geometry detect at most 17\% of the target interactions, also with the stronger reference (Section~\ref{sec:strongref}). A robot that uses such a threshold would treat most approaching people as passers-by and start its greeting or yielding behaviour too late; lowering the threshold would trade misses for false alarms, whose cost depends on how intrusive the triggered behaviour is \citep{riosmartinez2015proxemics,avelino2025greeting}. Calibrating the threshold on unlabelled or sparsely labelled data of the new site is therefore a precondition for deployment and was outside this study. Differences in class prevalence between development and deployment are known to invalidate thresholds and calibration \citep{godau2025navigating,silvafilho2023classifier}; calibration fitted across several source domains \citep{yu2022robust} and confidence thresholds estimated with unlabelled target data \citep{garg2022leveraging} are candidate remedies. The lead-time analysis is too small to show how early a robot can reliably act.

\subsection{The fixed reference as a measurement instrument}
Freezing the geometry branch did not yield better target rankings than joint training. What it provides is a statement about a model that exists: the claim that pose adds \DFGRev{} AP on HUI360 and no measurable amount on SSUP-A concerns a reference that can be deployed on its own and keeps its weights inside the combination, and the claim is paired for every seed and can be re-examined per date, lead time and control. For jointly trained models no such statement is possible, since their geometry term drifts. The construction also admits further cues without retraining earlier stages, and places no constraint on the reference beyond producing a logit, which the stronger-reference analysis exploited. When only the final ranking counts, the results give no reason to prefer FRPR over joint training or tree ensembles.

That the last-frame box area competes with every learned model follows partly from the benchmark: a positive window ends shortly before contact, when the person is necessarily close, and a negative window ends shortly before the moment of largest apparent size. Claims about pose or head orientation should therefore be made relative to a strong geometry baseline, which the paired design makes explicit.

\subsection{Limitations}
The labels are physical contacts with the platform; the models predict an event, not what a person intends. Windows are anchored on events, so the results describe ranking quality on a benchmark sample of tracks and do not give the false-alarm rate of a detector running continuously on a robot. The negative anchor is chosen with information from after the window, which aligns negatives with the moment of closest approach and may make apparent size more informative than it would be online. The official loader discards negatives with missing frames; with them restored, the main findings for FRPR hold, but the comparison with geometry-only tree ensembles in \fwd{} does not (Online Resource~1, Sect.~S6). On SSUP-A the azimuth channels describe the direction in the panorama rather than relative to the robot's interaction zone. Source-selected thresholds transferred poorly for all learned models. Both target sets had been used in earlier stages of this project, so none of the evaluations is blind; the orientation extension was locked before target scoring, whereas the stronger-reference analysis and the analysis of individual orientation cues are post hoc. The five and nine target recording dates bound the precision of every interval.

\section{Conclusion}\label{sec:conclusion}
We formulated interaction anticipation as a frozen geometry reference plus zero-initialised residuals on the logit, and used the exact split as an instrument to measure which cues transfer between two robots and sites. The pose correction raised AP over its reference by \DFGRev{} (95\% interval \CIFGRev) from SSUP-A to HUI360, with support from date, split, lead-time and pose-content controls, and gave no measurable gain in the opposite direction; the same asymmetry held over a stronger, source-selected reference. Geometry and pose kept their ranking quality in opposite transfer directions, a head-orientation residual added small gains in both, and whether a person faces the camera kept its meaning across datasets. Freezing did not improve AP over joint training, simple geometric baselines remained competitive, and source-selected thresholds transferred poorly. For social robots that must anticipate approach on a new platform, these results point to attention-related cues as candidates for transfer, to geometry and pose as cues whose value has to be re-established per platform, and to threshold calibration on the target site as a requirement before deployment.

\backmatter

\bmhead{Supplementary Information}
Online Resource~1 (ESM\_1.pdf) contains implementation details, the evaluation protocol, illustrations of the input construction and training procedure, the list of compared models, the sample partitions, per-seed results, the sensitivity analyses, selection results, all paired comparisons, the tree-ensemble, lead-time and sample-pool analyses, the orientation summary and cue screen, the stronger-reference protocol, and the case and attribution analysis.

\section*{Statements and Declarations}

\bmhead{Funding} This work was supported by the Jiangsu Provincial Major Science and Technology Program (No.~BG2025043), the National Natural Science Foundation of China (No.~52502402), the Basic Research Program of Jiangsu Province (No.~BK20251462) and the Natural Science Foundation of Jiangsu Province (No.~BK20232038).

\bmhead{Competing interests} The authors declare no competing interests.

\bmhead{Ethics approval and consent to participate} This study used only publicly released datasets, HUI360 and the SSUP-A annotation of the SSUP-HRI recordings \citep{lorenzolouis2026hui360,bu2024ssup}; the authors did not collect new data involving human participants. Not applicable.

\bmhead{Consent for publication} Not applicable. Fig.~1 and Figs.~S1 and S3 of Online Resource~1 show frames from the publicly released HUI360 dataset material \citep{lorenzolouis2026hui360}; no new images of individuals were collected.

\bmhead{Data availability} HUI360 and the SSUP-A annotations are distributed by their authors \citep{lorenzolouis2026hui360}; the SSUP-HRI recordings are described by \citet{bu2024ssup}. The core processed data, including the per-sample predictions and derived statistics behind every table and quantitative figure, are publicly available on GitHub at \url{https://github.com/WeiZhou96/FRPR-interaction-anticipation}.

\bmhead{Code availability} The core code is publicly available under the MIT licence on GitHub at \url{https://github.com/WeiZhou96/FRPR-interaction-anticipation}.

\bmhead{Author contributions} Bowen Yang: Conceptualization, Methodology, Software, Formal analysis, Investigation, Visualization, Writing -- original draft. Xinliang Xiao: Methodology, Validation, Formal analysis, Writing -- review and editing. Wenjing Zhang: Data curation, Validation, Visualization, Writing -- review and editing. Li Yang: Methodology, Supervision, Writing -- review and editing. Wei Zhou: Conceptualization, Supervision, Project administration, Funding acquisition, Writing -- review and editing. All authors read and approved the final manuscript.

\bmhead{Use of AI tools} Claude (Anthropic) was used to polish the English language of the manuscript. The authors reviewed and edited all text and take full responsibility for the content of the publication.

\bmhead{ORCID} Bowen Yang: 0009-0008-0229-0060; Xinliang Xiao: 0009-0001-9670-0575; Wenjing Zhang: 0009-0000-4262-0820; Li Yang: 0000-0002-8152-7642; Wei Zhou: 0000-0003-3225-0576.

\bibliography{references}

\end{document}